# COIL: Constrained Optimization in Learned Latent Space

Learning Representations for Valid Solutions

| Peter J. Bentley | Soo Ling Lim | Adam Gaier | Linh Tran |
|---|---|---|---|
| Dept. of Computer Science | Dept. of Computer Science | Autodesk Research | Autodesk Research |
| UCL, Autodesk Research | UCL | London, UK | London, UK |
| London, UK | London, UK | adam.gaier@autodesk.com | linh.tran@autodesk.com |
| p.bentley@cs.ucl.ac.uk | s.lim@cs.ucl.ac.uk | | |

## ABSTRACT

Constrained optimization problems can be difficult because their search spaces have properties not conducive to search, e.g., multimodality, discontinuities, or deception. To address such difficulties, considerable research has been performed on creating novel evolutionary algorithms or specialized genetic operators. However, if the representation that defined the search space could be altered such that it only permitted valid solutions that satisfied the constraints, the task of finding the optimal would be made more feasible without any need for specialized optimization algorithms. We propose *Constrained Optimization in Latent Space* (COIL), which uses a VAE to generate a learned latent representation from a dataset comprising samples from the valid region of the search space according to a constraint, thus enabling the optimizer to find the objective in the new space defined by the learned representation. Preliminary experiments show promise: compared to an identical GA using a standard representation that cannot meet the constraints or find fit solutions, COIL with its learned latent representation can perfectly satisfy different types of constraints while finding high-fitness solutions.

## KEYWORDS

Variational autoencoder, Learning latent representations, Genetic algorithm, Constrained optimization

## 1 INTRODUCTION

The combination of objective function, representation, and operators help define the space to be explored by any optimizer. For real-world problems, the objective function is typically limited by constraints [1]. Often the constraints can make a significant impact on the search space, adding discontinuities or conflicting with the objective function, which may result in ineffective optimization [2]. It is also common for such problems to require frequent re-optimization as constants may vary, while constraints and objectives remain the same (e.g., finding design alternatives [3, 4]).

Common solutions used in constrained optimization involve modifying the search operators or the optimization algorithm to overcome problems in the search space [5, 6]. These specialized algorithms may need tuning for each problem and expertise in constrained optimization, which may not always be available.

Recent work on learned representations for black-box optimization suggests an alternative approach [7, 8]. Instead of modifying the operators of the optimizer (or the optimizer itself), a new representation could be learned that is biased towards solutions that satisfy the constraints, such that discontinuities are minimized, and search becomes more feasible for the objective. Unlike existing work, these learned representations may not involve reduction of the number of parameters; instead, they would remap a hard-to-search genotype space into a more focused easier-to-search latent space, akin to the notion of evolution of evolvability [9] – except that Variational Autoencoders (VAEs) [10, 11] provide a faster method for learning the improved representation.

To this end, we propose *Constraint Optimization in Learned Latent Space* (COIL) which uses a VAE to generate a learned latent representation from a dataset comprising samples from the valid region of the search space according to a constraint, thus enabling the optimizer to find the objective in the new space defined by the learned representation. We provide an early exploration of this approach, examining improvements provided for constraint optimization in latent space vs. normal search space through two different types of constraint while using a relatively simple optimizer. We focus on the research question: *can a hard-to-search constrained space be mapped to an easier-to-search space?*

We also discuss the limitations of the approach and suggest improvements that may enable it to scale to more difficult, multiple constraints.

## 2 BACKGROUND

### 2.1 Constrained Optimization

The field of constrained optimization, sometimes referred to as constraint handling, focuses on the optimization of problems that comprise both an objective function and one or more constraints on the values that variables can take for that objective. The combination of objective function with constraints typically causes problems in real-world applications, as the constraints may not permit the true optimal, forcing the optimizer to compromise on the objective in order to meet the constraints [1].

Constraint handling methods have been studied for some time in the field of evolutionary computation, for example [2] and [12] described the typical early approach of using penalty values added to fitness scores when constraints are not met – an approach still commonly used in industry, while later work [6] provided an early classification of different constraint handling approaches that can be used in genetic algorithms.

Numerous competitions in the area have been held [13, 14], leading to the development of many specialized algorithms for constrained optimization. For example, [5] describes a widely-used tournament-selection-based approach where solutions that better satisfy constraints are chosen as winners in tournaments. In our work, COIL makes use of a similar tournament-based fitness calculation based on the same idea. A two-stage approach is proposed in [15], first using a GA to solve the constraints and then using nondominated search to find solutions that both satisfy the constraints and objective. In our work, COIL uses a similar two-stage approach. In [14], the authors analyze current approaches and provide a useful breakdown of different constraint types, such as the (non-)linearity of the constraints, the separability of the objective function and/or constraints, the relative size of the feasible region in the search space, the connectedness of the feasible region, and the orientation of the feasible region within the search space. In our work, we use some constraint types based on this analysis.

Despite the vast work in this domain, in the field of evolutionary computation, little work has been performed on the use of machine learning, and specifically Variational Autoencoders, to learn representations that correspond to valid areas of the search space according to constraints.

### 2.2 Variational Autoencoders

An autoencoder [16] is a class of neural networks that learns a representation of the data, typically used for feature learning or dimensionality reduction. Its extension, the Variational Autoencoder (VAE) [10] is a probabilistic autoencoder using variational inference. VAEs encode an observation into a latent space and reconstruct the observation based on the latent sample. VAEs have been widely used due to their simplicity (the latent space is set to be an isotropic Gaussian distribution) and the resulting analytical solution for the regularization.

### 2.3 Evolving Latent Variables

Combinations of deep learning approaches for representation learning and evolutionary approaches have had many recent successes. This technique of latent variable evolution (LVE) uses generative models to learn low dimensional real-valued representations of existing datasets, and then search with evolution within that compressed space. LVE approaches have been used to search within the space of fingerprints to defeat security [17], the space of Mario levels for new levels with different levels of enemies [18], celebrity faces with varied colors of hair and eyes [19] and human portrait generation [20].

If no existing datasets are available, they must be generated. Generating these dataset can be accomplished by running on optimizer on a problem many times, each time adding the best solution to a training set [7, 21]. A representation learned from this set of solutions encodes a bias which can be useful for solving similar kinds of problems.

Dataset generation can also be accomplished with Quality-Diversity approaches [22, 23], which generate collections of diverse high performing solutions. These solutions can form a dataset to train generative models, in some cases using those generative models within the same optimization run to generate ever high quality solutions and encodings [8],[24]. Even standard and surrogate-assisted GAs can benefit from learning encodings during optimization to better tackle higher-dimensional search spaces [25, 26].

Representations learned from high performing solutions reduce the range of solutions which can be found [27]. Though problematic when a model trained only on white faces will not produce one of color, a learned encoding that can only produce high performing solutions can simplify search immensely.

## 3 METHOD

Our proposed COIL approach comprises three steps:
Step 1: Generation of Valid Data; Step 2: Representation Learning, and Step 3: Optimization.

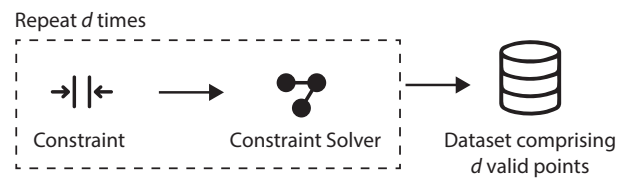

**Figure 1: Generating valid data that satisfy the constraint.**

The first step of COIL (Figure 1) is to create a dataset corresponding to solutions that satisfy the constraint alone. This is considerably easier than attempting to find an optimal solution that also satisfies the constraint. We use the simple genetic algorithm of the DEAP framework [27] with real encoding. Fitness is defined only by the constraint, with individuals worse at satisfying the constraint given a worse fitness and individuals that achieve a fitness of zero (i.e., that meet the constraint) being added to the dataset. A dataset of $d$ values is generated by running the GA multiple times. The initially random populations help enable an even random sampling of feasible values. While here we have used a simple GA, alternative algorithms could provide more efficient or more effective coverage, e.g. Clustering [28], Clearing [29], Novelty Search, or MAP-Elites [22, 23, 30] or specialized constraint satisfaction algorithms for more challenging constraints [31, 32].

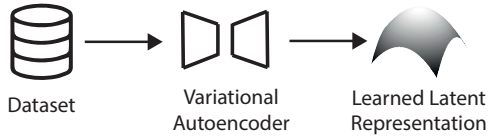

**Figure 2: Learning a latent representation biased towards solutions that satisfy the constraint.**

For the second step (Figure 2) we use a simple VAE[1] with loss function from [10] to learn a latent representation from the dataset, learning for $E$ epochs. We do not perform parameter reduction – here we are mapping the original space to a new, easier-to-search space with the same number of variables.

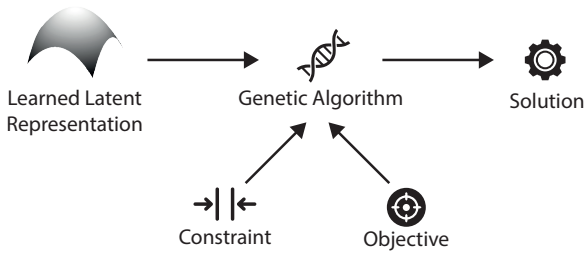

**Figure 3: Optimizing over the latent representation to find solutions that both satisfy the constraint and meet the objective.**

The third step (Figure 3) makes use of this learned representation. We use a simple genetic algorithm (same settings as in step 1) with the objective function to search in the learned latent space. The learned VAE model is used to map the evolving latent variables to actual problem variables and fitness is the objective and the original constraint used to generate the dataset in the first step. We use a tournament fitness measure [5] (Algorithm 1) to ensure objective and constraint are treated equally while removing any need to sum and weight or use penalty values. Because the learned latent representation can only represent valid solutions, the GA functions well with a small population size evolving for very few generations.

---
[1] https://github.com/pytorch/examples/tree/master/vae

**Algorithm 1: Tournament fitness**

reset fitness of all individuals to 0
for $i$ = 1 to *popsize*
   competitiongroup = 5 randomly picked individuals
   for $j$ = 1 to 10
     pick 2 random individuals ($a$, $b$) from competitiongroup
     $a_{num\_match}$++
     $b_{num\_match}$++
     for $e$ = 1 to *num_criteria*
       if fit$_e(a)$ is better than fit$_e(b)$ then $a_{fitness}$++
       if fit$_e(b)$ is better than fit$_e(a)$ then $b_{fitness}$++
       elseif fit$_e(b)$ = fit$_e(a)$ then $a_{fitness}$++ and $b_{fitness}$++
     endfor
   endfor
endfor
for every individual $i$ in population
   if ($i_{num\_match}$ > 0) $i_{fitness}$ = $i_{fitness}$ / $i_{num\_match}$
   else $i_{fitness}$ = 0
endfor

## 4 PRELIMINARY EXPERIMENTS

For all experiments we use a simple objective function to minimize for $D$ variables:

$$f(\overline{x}) = \sum_{i=0}^{D-1} x_i^2$$

We examine two simple constraints in these preliminary experiments, which represent commonly observed constraint types: the simple bound constraint (C1):

$$\sum_{i=0}^{D-1}(45 - x_i) \leq 0$$

and correlated variables in the form of chained inequality (C2):

$$\forall i \in \{0,\ldots, D-2\}\ \ 8 \leq (x_{i+1} - x_i) \leq 10$$

C1 and C2 transform the objective into a task that cannot be solved reliably using a standard GA by disallowing the true optimal, forcing the optimizer to compromise on the objective to meet the constraint (Figure 4). While C1 can be readily solved on its own by a GA, C2 is more difficult and so for the data generation stage for this constraint we use a specific C2 solver (Algorithm 2). We apply each step of COIL: generating data, learning new representation, using a GA to find optimal solutions with this representation. First, we examine the VAE and its learned representation.

**Algorithm 2: C2 Constraint Solver and Data Generator**

datapoint = [];
# bounds:
XMIN = -50; XMAX = 50; B1 = 8.0; B2 = 10.0
# start with first pair or triplet of points
$r$ = random.uniform(XMIN, XMAX)
if the number of variables $D$ is even
   r1 = r + random.uniform(B1, B2)
   datapoint = [r, r1]
   count = 2
else: # odd
   r0 = r - random.uniform(B1, B2)
   r1 = r + random.uniform(B1, B2)
   datapoint = [r0, r, r1]
   count = 3
# generate remaining points, 2 at a time
while count < $D$
   # prepend val - rand() to the list
   datapoint = [datapoint[0] - random.uniform(B1, B2)] + datapoint
   # append val + rand() to the list
   datapoint.append(datapoint[-1] + random.uniform(B1, B2))
     count += 2
if (datapoint[0] >= XMIN) and (datapoint[-1] <= XMAX):
  return unnormalise_to_range(datapoint)
else:
  return None

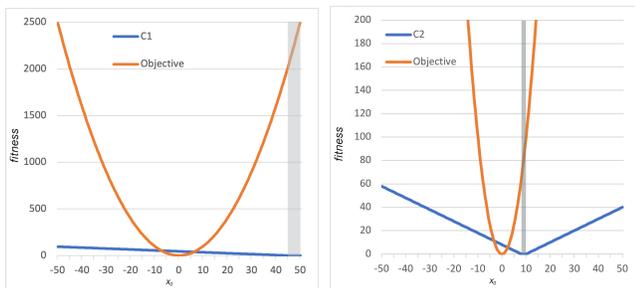

**Figure 4: Plots of $x_0$ against fitness for C1 and C2, $D$=2 at $x_1$=0 (blue) and the objective function (orange), where 0 is optimal for all functions. Grey regions indicate valid regions of search space according to the constraints. In order to satisfy the constraints, it is necessary to find the best valid solution for the objective, which is some distance from the unconstrained optimal. The difficulty increases as the number of variables $D$ increases, especially for C2 as a chain of dependencies is created.**

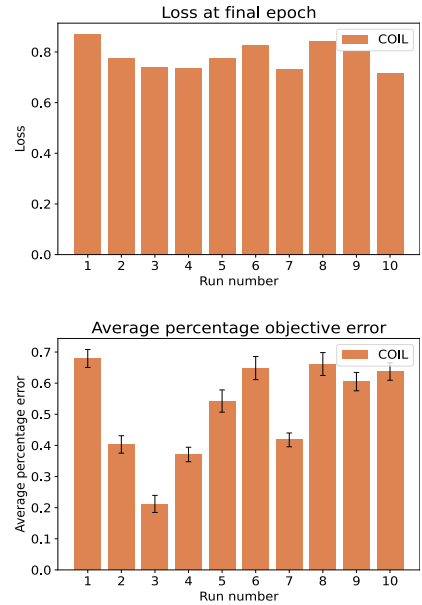

**Figure 5: Loss at final epoch of VAE for 10 different runs and corresponding average error of optimizing objective with C1 over 100 runs. Error bars: mean ± SE.**

## 4.1 Assessing Reliability of Learned Representation

We generated a dataset for C1 with $D = 3$ and then ran Step 2 (Representation Learning) 10 times on the dataset, to assess loss variability at the final epoch and objective error when optimizing with the corresponding learned representation. Results show variability in the success of the VAE, with a clear correspondence between smaller VAE losses and better GA optimization results (Figure 5). VAEs with worse losses provide worse reconstruction, i.e., the result of encoding and then decoding values using the learned model may not resemble the input. Thus, a GA that uses a poorly learned latent representation will have an inconsistent mapping from latent variable to actual variable, making evolution more difficult. Based on this, our default Representation Learning Step 2 runs 10 times and uses the VAE model with the lowest loss.

## 4.2 Comparing Learned Latent Representation with Standard Representation

We applied COIL to C1 and C2, varying the problem size (number of variables $D$ for constraints and objective) from 1 to 10. For data generation the GA population size was 200, evolving for a maximum of 200 generations, repeated to make a dataset of 5000. Termination criteria comprises fitness achieving zero or maximum generations being reached. The VAE used 4 linear layers, learning for 200 epochs, with KLD = 1, learning rate 0.001 using Adam as the optimizer. The GA used to evolve the learned latent representation needed a population size of just 20 for 50 generations.

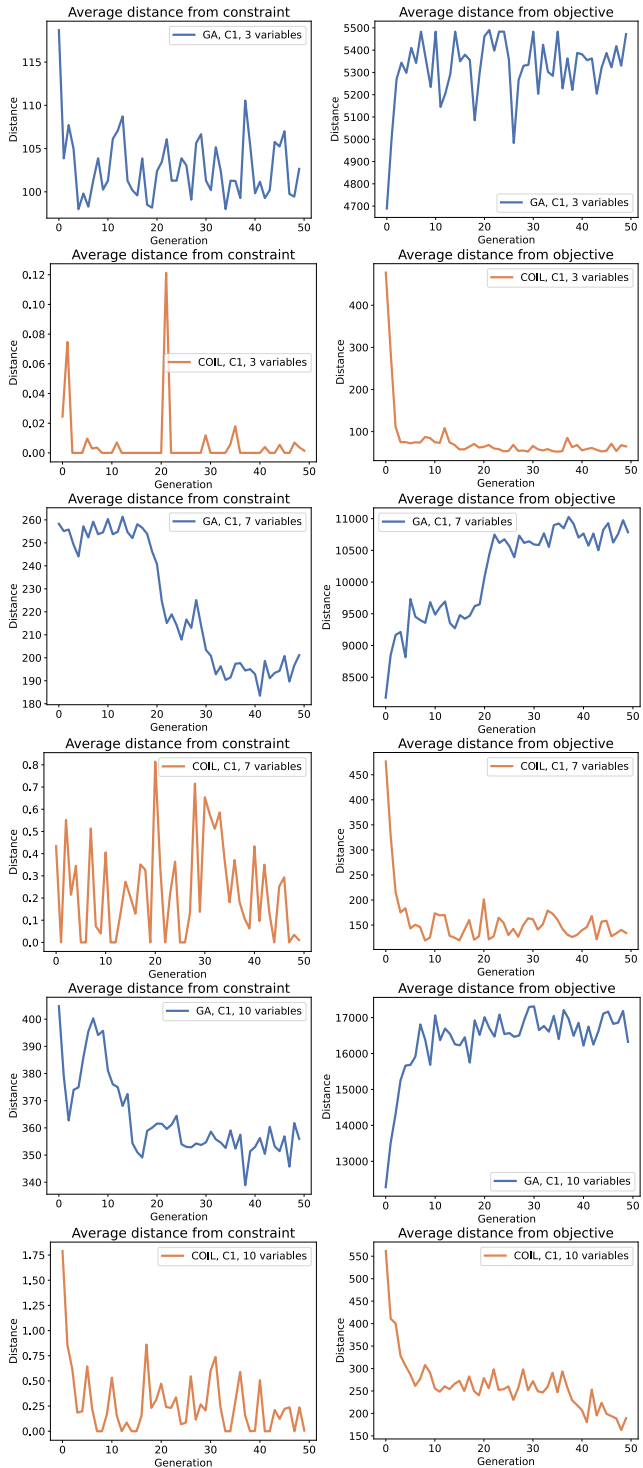

Figure 6: Example run showing COIL C1 constraint error, standalone GA C1 constraint error, COIL objective error over time, and standalone GA objective error over time for $D = 3$, 7 and 10 variables, and averaged over entire population. Note difference in y-axis scales for COIL and standalone GA results.

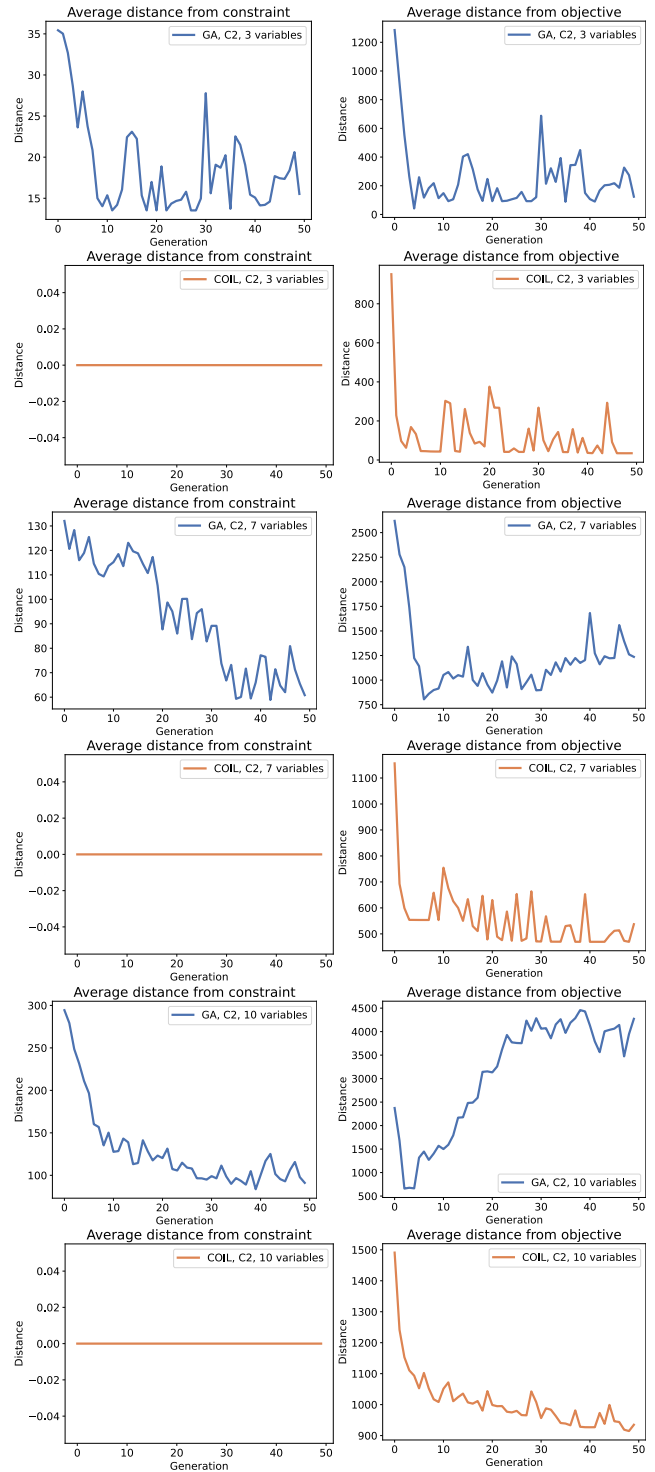

Figure 7: Example run showing COIL C2 constraint error, standalone GA C2 constraint error, COIL objective error over time, and standalone GA objective error over time for $D = 3$, 7 and 10 variables, and averaged over entire population. Note difference in y-axis scales for COIL and standalone GA results.

For both GAs, uniform crossover was used with probability 0.05 per individual, Gaussian mutation with probability 0.2 per individual, mu=0, sigma=1, tournament selection (tournament size of 3) is used, values of genes are bound to between -1.0 and 1.0 for initialization and for all operators.

COIL was compared with a standalone GA evolving a direct representation of the same problem (with identical settings). Every run for both representations was repeated 100 times.

Figure 6 and Figure 7 shows representative example runs for constraint C1 and C2 respectively, for 3, 7 and 10 variables. Overall, the GA performed poorly for C1 and C2, with poor results according to the objective function. COIL was able to achieve near perfect results for C1 and C2 with fit solutions according to the objective.

When examining the progress of evolution over time it is apparent that the standalone GA struggles – the combination of the constraint and objective has transformed the search space such that the standalone GA cannot search effectively. For C1, the standalone GA sacrifices good objective scores as it attempts to find valid solutions, resulting in poor results for both. For C2, the standalone GA sometimes finds reasonable solutions for the objective, but such scores are "cheating" as the constraint is not satisfied correctly.

In contrast, COIL solutions during evolution are always very near to perfect for C1 and always perfect for C2 in terms of constraint satisfaction. When compared to the solutions generated by the standalone GA, this clearly indicates that the learned latent representation produces a useful bias towards the representation of valid solutions. Similarly, when examining the quality of solutions in terms of the objective during evolution, COIL shows a consistent and rapid improvement towards the best possible solution in every case while maintaining constraint satisfaction for all variables in the problem. All solutions achieve better scores for the overall objective compared to the standalone GA, often by a huge margin. The results are a positive indication that the representation is evolvable for the GA – it can modify the latent variables as it normally would modify its parameters (genes) and traverse the latent space successfully.

Figure 8 visualizes the latent space by varying the learned latent variables for C1 and C2, with $D$=1 and $D$=2 respectively. In this example, the VAE has clearly learned a representation that always satisfies the constraint C1, as all possible values for the latent variable map onto valid solutions between 45 and 50. The VAE also remapped the space for C2, learning the dependency between the problem variables, and using the first latent variable to set the overall value for both expressed values while keeping them separated by approximately 8 as required C2, again with the result that constraint C2 will always be satisfied. The second latent variable also adjusts both variables, but only up to 0.4.

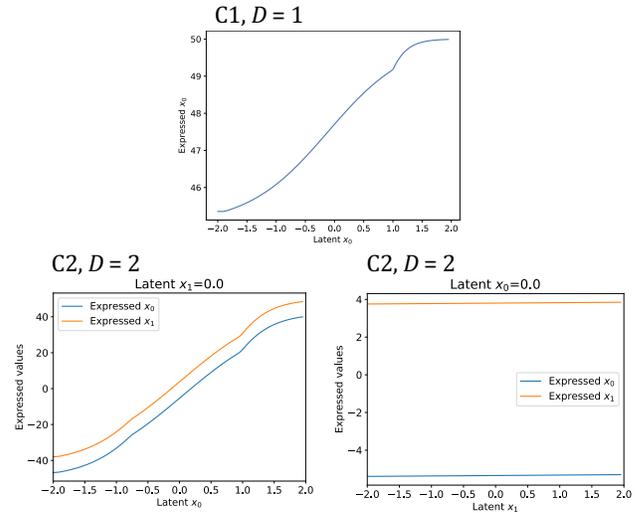

Figure 8: Visualizing latent variables. Top: varying latent variable produces output only between 45 and 50, thus satisfying constraint C1. Bottom: varying $x_0$ with $x_1$ fixed produces a full range of expressed values for both variables with a gap of 8 at all times, meeting constraint C2. Varying $x_1$ with $x_0$ fixed produces a change of up to 0.4 for both variables, enabling fine-tuning.

While this variable may be considered superfluous, for the GA this is a useful variable for fine-tuning the result when searching in the space for the optimal. Both learned latent spaces provide sensible approaches to overcome the constraints and a researcher experienced in designing representations suitable for a GA would likely make similar choices. The VAE has enabled COIL to learn its own representations, ensuring constraints are satisfied.

## 5 DISCUSSION

### 5.1 Assessing Overhead of Approach

Once the latent representation has been learned, because of the speed of using a pretrained VAE model, there is minimal overhead between evolving latent representations vs. standard, for example 100 runs of the 3 variable C1 takes 8.43 seconds evolving the latent representation vs. 3.74 seconds evolving a standard representation (no parallelism used). With appropriate parallelization the differences will be negligible.

The main overhead comes from the valid data-generation step. Clearly the creation of 5000 valid solutions per constraint takes time and may make COIL slower than state-of-the-art constrained optimizers to run. The approach used in this work has not been optimized and the use of parallelization and techniques such as Quality diversity [22] would enable more efficient creation of datasets. However, just as contemporary deep learning approaches have reduced the need for lengthy feature engineering in machine learning during the implementation stages, COIL has the potential to reduce the need for lengthy representation, search operator and algorithm design, enabling simple optimizers to solve problems that they otherwise could not. COIL is also suited for real-world problems where constraints remain fixed, but objectives/constants vary and re-optimization is frequent. COIL shows how we can solve the part of the problem defined by the constraints once and incorporate that knowledge into the representation, saving time for every future instance of optimization that occurs.

### 5.2 Novelty of Solutions Found using Learned Representation

Step one of this approach involves producing datasets of valid solutions that satisfy the constraint(s). Should the optimal solution to the objective also appear in these datasets – discovered by chance in the random sampling of the valid space – it could be argued that the VAE and final GA are redundant. This is most likely to occur for simple constraints, so we analyze C1 datasets. For a single variable, similar but non-identical solutions to the optimal did appear in the dataset. For two or more variables, no examples of optimal solutions appeared in the datasets – COIL created novel solutions using the learned representation. For more complex constraints and multiple constraints it is highly unlikely that random sampling will discover valid optimal solutions and thus appear in the training sets, otherwise such problems would be readily solvable using random search, which they are not.

### 5.3 Reuse of Learned Representation

One significant advantage of COIL can be seen when multiple independent variables are each limited by the same constraint. Applying the representation learned on one variable to all similarly constrained problem parameters enables massive scalability – the same learned model could be used on an unlimited number of such problem parameters.

### 5.4 Tackling Multiple Constraints and Discontinuous Functions

The simple method proposed here requires extension to handle more complex problems. When the problem comprises multiple constraints, each a function of a subset of the problem variables, then it is no longer possible to construct a single dataset for the VAE. Instead, we can create a "Stacked COIL" where each constraint is considered sequentially, and the learned representation for one is used to generate a dataset for the next, and so on (Figure 9).

Conversely, if the constraint is a discontinuous function, our simple VAE will not be able to learn this distribution. Its assumption that the input fits a Gaussian distribution prevents it from learning more complex distributions.

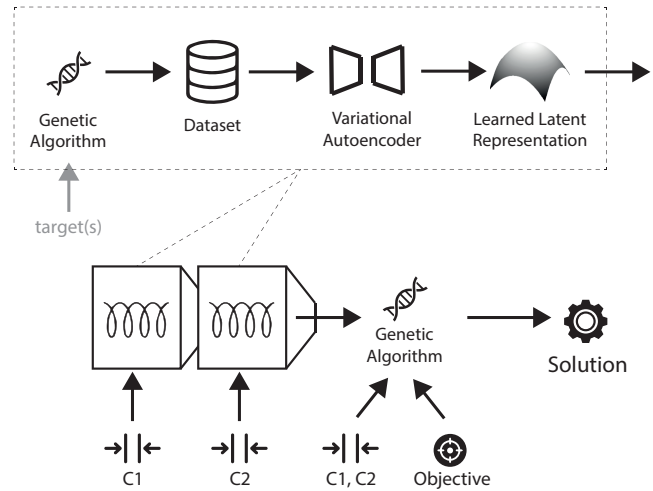

**Figure 9: Stacked COIL concept, enabling the incremental learning of a latent space that is biased towards valid solutions for multiple constraints applied to different subsets of problem variables.**

Extending the VAE to a mixture model, or using a normalizing flow prior overcomes this limitation. [33] provides details of our ongoing work on these extensions, broadening the idea to multiple criteria and constraints.

### 5.5 Constraint Handling Limitations

When evolving its learned latent variables, COIL works by treating constraints as additional problem criteria, awarding a fitness score based on the degree to which a constraint is satisfied. While this can be an effective way of handling soft constraints, this does not guarantee any constraint will be satisfied unless the VAE has successfully learned a representation that can only express valid solutions – also not guaranteed. For this reason, COIL may not be suitable for hard constraints where every constraint must be satisfied perfectly. COIL is not a conventional constraint satisfaction approach – it is a method for transforming heavily constrained search spaces.

## 6 CONCLUSIONS

Constraints can distort a search space into a tangle of no-go areas, discontinuities, and dependencies. Here we have proposed Constraint Optimization in Learned Latent Space (COIL), which uses representation learning in order to transform the original search space into a space easier to search by learning a bias towards valid solutions. We have demonstrated that this approach can achieve substantive improvements compared to a standalone optimizer for different problem sizes. Further work continues this theme, extending the idea of COIL to additional problem criteria [33].

While this work has used a GA for data generation and optimization and a VAE for representation learning, COIL can use any equivalent approaches, e.g., a constraint solver, generative adversarial learning, and gradient descent. Future work should investigate the most appropriate methods for each stage, given different types of optimization problem. Given the advances in scalability and efficiency of generative machine learning, we anticipate that this combination of ML with optimization has the potential to transform our ability to perform complex real-world optimization.

## SOURCE CODE

The source code necessary to reproduce the experiments in this paper is available at:

https://github.com/writingpeter/coil_gecco22